\documentclass[journal]{IEEEtran}
\usepackage{graphicx}
\usepackage{subfigure}
\usepackage{multirow}
\usepackage{booktabs}
\usepackage[numbers]{natbib}
\usepackage{amsmath}
\usepackage{lineno}
\usepackage{amsmath}

\ifCLASSINFOpdf
\else
\fi
\begin{document}
%
\title{Hierarchical Multi-scale Attention Networks for Action Recognition}
%
%
%

\author{Shiyang~Yan,
        Jeremy~S. Smith,
        Wenjin~Lu 
        and~Bailing~Zhang
\thanks{Shiyang Yan, Wenjin Lu and Bailing Zhang are with the Department
of Computer Science and Software Engineering, Xi'an Jiaotong-liverpool University, Suzhou, China.}
\thanks{Jeremy S. Smith is with the University of Liverpool.}
}
\maketitle

\begin{abstract}
Recurrent Neural Networks (RNNs) have been widely used in natural language processing and computer vision. Among them, the Hierarchical Multi-scale RNN (HM-RNN), a kind of multi-scale hierarchical RNN proposed recently, can learn the hierarchical temporal structure from data automatically. In this paper, we extend the work to solve the computer vision task of action recognition. However, in sequence-to-sequence models like RNN, it is normally very hard to discover the relationships between inputs and outputs given static inputs. As a solution, attention mechanism could be applied to extract the relevant information from input thus facilitating the modeling of input-output relationships. Based on these considerations, we propose a novel attention network, namely Hierarchical Multi-scale Attention Network (HM-AN), by combining the HM-RNN and the attention mechanism and apply it to action recognition. A newly proposed gradient estimation method for stochastic neurons, namely Gumbel-softmax, is exploited to implement the temporal boundary detectors and the stochastic hard attention mechanism. To amealiate the negative effect of sensitive temperature of the Gumbel-softmax, an adaptive temperature training method is applied to better the system performance. The experimental results demonstrate the improved effect of HM-AN over LSTM with attention on the vision task. Through visualization of what have been learnt by the networks, it can be observed that both the attention regions of images and the hierarchical temporal structure can be captured by HM-AN.

\end{abstract}


%
\IEEEpeerreviewmaketitle

\section{Introduction}
Action recognition in videos is a fundamental task in computer vision. Recently, with the rapid development of deep learning, and in particular, deep convolutional neural networks (CNNs), a number of models \cite{krizhevsky2012imagenet} \cite{simonyan2014very} \cite{szegedy2015going} \cite{7780459} have been proposed for image recognition. However, for video-based action recognition, a model should accept inputs with variable length and generate the corresponding outputs. This special requirement makes the conventional CNN model that caters for one-versus-all classification unsuitable.

RNNs have long been explored for sequential applications for decades, often with good results. However, a significant limitation of vanilla RNN models which strictly integrate state information over time is the ¡°vanishing gradient¡± effect \cite{bengio1994learning}: the ability to back propagate an error signal through a long-range temporal interval becomes increasingly impossible in practice. To mitigate this problem, a class of models with a long-range dependencies learning capability, called Long Short-Term Memory (LSTM), was introduced by Hochreiter and Schmidhuber \cite{hochreiter1997long}. Specifically, LSTM consists of memory cells, with each cell containing units to learn when to forget previous hidden states and when to update hidden states with new information.

Many sequential data often have complex temporal structure which requires both hierarchical and multi-scale information to be modeled properly. In language modeling, a long sentence is often composed of many phrases which further can be decomposed by words. Meanwhile, in action recognition, an action category can be described by many sub-actions. For instance, `long jump' contains `running', `jumping' and `landing'. As stated in \cite{chung2016hierarchical}, a promising approach to model such hierarchical representation is the multi-scale RNN. One popular approach of implementing multi-scale RNN is to treat the hierarchical timescales as pre-defined parameters. For example, Wang et al. \cite{wang2016hierarchical} implemented a multi-scale architecture by building a multiple layers LSTM in which higher layer skip several time steps. In their paper, the skipped number of time steps is the parameter to be pre-defined. However, it is often impractical to pre-define such timescales without learning, which also leads to poor generalization capability. Chung et al. \cite{chung2016hierarchical} proposed a novel RNN structure, Hierarchical Multi-scale Recurrent Neural Network (HM-RNN), to learn time boundaries from data automatically. These temporal boundaries are like rules described by discrete variables inside RNN cells. Normally, it is difficult to implement training algorithms for discrete variables.
Popular approaches include unbiased estimator with the aid of REINFORCE \cite{bengio2013estimating}. In this paper, we re-implement the HM-RNN by applying a recently proposed Gumbel-sigmoid function \cite{jang2016categorical} \cite{DBLP:journals/corr/MaddisonMT16} to realize the training of stochastic neurons due to its efficiency in practice \cite{gulcehre2017memory}.

In the general RNN framework for sequence-to-sequence problems, the information of input is treated uniformly without discrimination on the different parts. This will result in fixed length of intermediate features and subsequent sub-optimal system performance. On the other hand, the practice is in sharp contrast to the way humans accomplish sequence processing tasks. Humans tend to selectively concentrate on a part of information and at the same time ignores other perceivable information. The mechanism of selectively focusing on relevant contents in the representation is called attention. The attention based RNN model in machine learning was successfully applied in natural language processing (NLP), and more specifically, in neural translation \cite{bahdanau2014neural}. For many visual recognition tasks, different portions of an image or segments of a video have unequal importance, which should be selectively weighted with attention. Xu et al. \cite{xu2015show} systematically analyzed the stochastic hard attention and deterministic soft attention models and applied them in image captioning tasks, with improved results compared with other RNN-like algorithms. The hard attention mechanism requires a stochastic neuron which is hard to train using conventional back propagation algorithm. They applied REINFORCE \cite{bengio2013estimating} as an estimator to implement hard attention for image captioning.

The REINFORCE is an unbias gradient estimator for stochastic units, however, it is very complex to implement and often with high gradient variance during training \cite{gulcehre2017memory}. In this paper, we study the applicability of Gumbel-softmax \cite{jang2016categorical} \cite{DBLP:journals/corr/MaddisonMT16} in hard attention because Gumbel-softmax is an efficient way to estimate discrete units during training of neural networks. To mitigate the problem of sensitive temperature in Gumbel-softmax, we apply an adaptive temperature scheme \cite{gulcehre2017memory} in which the temperature value is also learnt from data. The experimental results verify that the adaptive temperature is a convenient way to avoid manual searching for the parameter. Additionally, we also test the deterministic soft attention \cite{xu2015show} \cite{sharma2015action} and stochastic hard attention implemented by REINFORCE-like algorithms \cite{mnih2014recurrent} \cite{ba2014multiple} \cite{xu2015show} in action recognition. Combined with HM-RNN and two types of attention models, we systematically evaluate the proposed Hierarchical Multi-scale Attention Networks (HM-AN) for action recognition in videos, with improved results.

Our main contributions can be summarized as follows:

\begin{itemize}
  \item We propose a Hierarchical Multi-scale Attention Network (HM-AN) by combining HM-RNN and both stochastic hard attention and deterministic soft attention mechanism for vision tasks.
  \item By re-implementing Gumbel-softmax and Gumbel-sigmoid, we make the stochastic neurons in the networks trainable by back propagation.
  \item We test the proposed model on action recognition from video, with improved results.
  \item Through visualization of the learnt attention regions and boundary detectors of HM-AN, we provide insights for further research.
\end{itemize}

\section{Related Works}

\subsection{Hierarchical RNNs}
The modeling of hierarchical temporal information has long been an important topic in many research areas. The most notable model is LSTM proposed by Hochreiter and Schmidhuber \cite{hochreiter1997long}. LSTM employs the multi-scale updating concept, where the hidden units' update can be controlled by gating such as input gates or forget gates. This mechanism enables the LSTM to deal long term dependencies in temporal domain. Despite the advantage, the maximum time steps are limited within a few hundreds because of the leaky integration which makes the memory for long-term gradually diluted \cite{chung2016hierarchical}. Actually, the maximum time steps in video processing is several dozens which makes the application of LSTM in video recognition very challenging.

To alleviate the problem, many researchers tried to build a hierarchical structure explicitly, for instance, Hierarchical Attention Networks (HAN) proposed in \cite{wang2016hierarchical}, which is implemented by skipping several time steps in the higher layers of the stacked multi-layer LSTMs. However, the number of time steps to be skipped is a pre-defined parameter. How to choose these parameters and why to choose certain number are unclear.

More recent models like clockwork RNN \cite{koutnik2014clockwork} partitioned the hidden states of RNN into several modules with different timescales assigned to them. The clockwork RNN is more computationally efficient than the standard RNN as the hidden states are updated only at the assigned time steps. However, finding the suitable timescales is challenging which makes the model less applicable.

To mitigate the problem, Chung et al. \cite{chung2016hierarchical} proposed Hierarchical Multiscale Recurrent Neural Network (HM-RNN). HM-RNN is able to learn the temporal boundaries from data, which facilitates the RNN model to build a hierarchical structure and enable long-term dependencies automatically. However, the temporal boundaries are stochastic discrete variables which are very hard to train using standard back propagation algorithm.

A popular approach to train the discrete neurons is the REINFORCE-like \cite{williams1992simple} algorithms. It is an unbiased estimator but often with high gradient variance \cite{chung2016hierarchical}. The original HM-RNN applied straight-through estimator \cite{bengio2013estimating} because of its efficiency and simplicity in practice. In this paper, instead, we applied a more recent Gumbel-sigmoid \cite{jang2016categorical} \cite{DBLP:journals/corr/MaddisonMT16} to estimate the stochastic neurons. It is much more efficient than other approaches and achieved state-of-the-art performance among many other gradient estimators \cite{jang2016categorical}.

\subsection{Attention Mechanism}

One important property of human perception is that we do not tend to process a whole scene in its entirety at once. Instead humans pay attention selectively on parts of the visual scene to acquire information where it is needed \cite{mnih2014recurrent}. Different attention models have been proposed and applied in object recognition and machine translation. Mnih et al. \cite{mnih2014recurrent} proposed an attention mechanism to represent static images, videos or as an agent that interacts with a dynamic visual environment. Also, Ba et al. \cite{ba2014multiple} presented an attention-based model to recognize multiple objects in images. The two above-mentioned models are all with the aid of REINFORCE-like algorithms.

The soft attention model was proposed for machine translation problem in NLP \cite{bahdanau2014neural}, and Xu et al. \cite{xu2015show} extended it to image caption generation as the task is analogous to `translating' an image to a sentence. Specifically, they built a stochastic hard attention model with the aid of REINFORCE and a deterministic soft attention model. The two attention mechanisms were applied in image captioning task, with good results. Subsequently, Sharma et al. \cite{sharma2015action} built a similar model with the soft attention applied in action recognition from videos.

There is a number of subsequent works on the attention mechanism. For instance, in \cite{yao2015video}, the attention model is utilized for video description generation by softly weighting the visual features extracted from frames in each video. Li et al. \cite{li2016videolstm} combined convolutional LSTM \cite{xingjian2015convolutional} with soft attention mechanism for video action recognition and detection. Teh et al. \cite{teh2016attention} extended the soft attention into CNN networks for weakly supervised object detection.

One important reason for applying soft attention instead of its hard version is that the stochastic hard attention mechanism is difficult to train. Although the REINFORCE-like algorithms \cite{williams1992simple} are unbiased estimators to train stochastic units, their gradients have high variants. To solve the problem, recently, Jang et al. \cite{jang2016categorical} proposed a novel categorical re-parameterization techniques using Gumbel-softmax distribution. The Gumbel-softmax is a superior estimator for categorical discrete units \cite{jang2016categorical}. It has been proved to be efficient and has high performance in \cite{jang2016categorical}.

\subsection{Action Recognition}
RNNs have been popular for speech recognition \cite{graves2013hybrid}, image caption generation \cite{xu2015show}, and video description generation \cite{yao2015video}. There have also been efforts made for the application of LSTM RNNs in action recognition. For instance, \cite{donahue2015long} proposed an end-to-end training system using CNN and RNNs deep both in space and time to recognize activities in video. \cite{yue2015beyond} also explicitly models the video as an ordered sequence of frames using LSTM. Most of the previous works treat image features extracted from CNNs as static inputs to RNNs to generate action labels at each frame. However, static features indiscriminating of important and irrelevant information will be detrimental to system performance. On the other hand, the interpretation of CNN features will be much easier if attention model can be applied for action recognition because attention mechanism automatically focuses on specific regions to facilitate the classification.

In this paper, as discussed previously, we re-implement the HM-RNN to capture the hierarchical structure of temporal information from video frames. By incorporating the HM-RNN with both stochastic hard attention and deterministic soft attention, long-term dependencies of video frames can be captured.

Works related to ours also include the attention model proposed by Xu et al. \cite{xu2015show} and \cite{jin2015aligning}. \cite{xu2015show} first applied both stochastic hard attention and deterministic soft attention mechanisms for spatial locations of images for image captioning. \cite{jin2015aligning} instead used weighting on image patches to implement a region-level attention. In this paper, similar to \cite{xu2015show}, both of stochastic hard attention and deterministic soft attention are studied. However, when implementing hard attention, \cite{xu2015show} borrowed the idea of REINFORCE whilst we also propose to apply a more recent Gumbel-softmax to estimate discrete neurons in the attention mechanism.

\section{The proposed methods}
In this section, we first re-visit the HM-RNN structure proposed in \cite{chung2016hierarchical}, then introduce the proposed HM-AN networks, with details of Gumbel-softmax and Gumbel-sigmoid to estimate the stochastic discrete neurons in the networks.
\begin{figure*}[t]
  \centering
  \includegraphics[width=\textwidth]{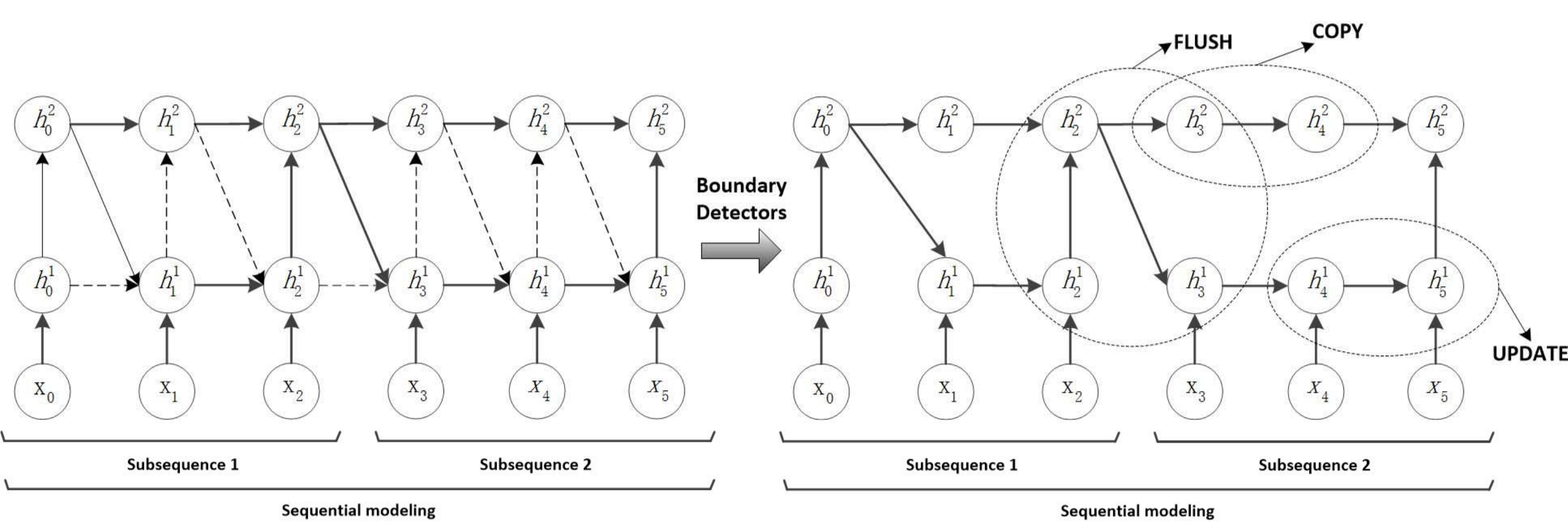}\\
  \caption{Network Structure: After the networks discover the implicit boundary relations of the multi-scale property, boundary detectors can set the networks into an explicit multi-scale architecture. }\label{system}
\end{figure*}

\subsection{HM-RNN}

HM-RNN was proposed in \cite{chung2016hierarchical} to better capture the hierarchical multi-scale temporal structure in sequence modeling. HM-RNN defines three operations depending on the boundary detectors: UPDATE, COPY and FLUSH. The selection of these operations is determined by boundary state $z_t^{l-1}$ and $z_{t-1}^l$, where $l$ and $t$ stand for the current layer and time step, respectively:

\begin{equation}\label{boundary}
  \begin{array}{ll}
    UPDATE, & \hbox{$z_{t-1}^l = 0$ and $z_t^{l-1} = 1$;} \\
    COPY, & \hbox{$z_{t-1}^l = 0 $ and $z_t^{l-1} = 0$;} \\
    FLUSH, & \hbox{$z_{t-1}^l = 1$.}
  \end{array}
\end{equation}

The updating rules of operation UPDATE, COPY and FLUSH are defined as follows:

\begin{equation}\label{cell}
 c_{t}^{l} = \begin{cases}
f_{t}^{l} \odot c_{t-1}^{l} + i_{t}^{l} \odot g_{t}^{l},   &  UPDATE\\
c_{t-1}^{l}, & COPY \\
i_{t}^{l} \odot g_{t}^{l},  & FLUSH
\end{cases}
\end{equation}

The updating rules for hidden states are also determined by the pre-defined operations:

\begin{equation}\label{hidden}
 h_{t}^{l} = \begin{cases}
 h_{t-1}^{l},   &  COPY\\
 o_{t}^{l} \odot c_{t}^{l}, & otherwise \\
\end{cases}
\end{equation}

The (i, f, o) indicate input, forget and output gate, respectively. g is called `cell proposal' vector. One of the advantages of HM-RNN is that the updating operation (UPDATE) is only executed at certain time steps sparsely instead of all, which largely reduce the computation cost.

The COPY operation simply copies the cell memory and hidden state from previous time step to current time step in upper layers until the end of a subsequence, as shown in Fig. \ref{system}. Hence, upper layer is able to capture coarser temporal information. Also, the boundaries of subsequence are learnt from data which is a big improvement of other related models. To start a new subsequence, FLUSH operation needs to be executed. FLUSH operation firstly forces the summarized information from lower layers to be merged with upper layers, then re-initialize the cell memories for the next subsequence.

In summary, the COPY and UPDATE operations enable the upper and lower layers to capture information on different time scales, thus realizing a multi-scale and hierarchical structure for a single subsequence. On the other hand, FLUSH operation is able to summarize the information from last subsequence and forward them to next subsequence, which guarantee the connection and coherence between parts within a long sequence.

The values of gates (i, f, o, g) and boundary detector z are obtained by:

\begin{equation}
\label{update}
{\left(
\begin{aligned}
i_t^l\\
f_t^l\\
o_t^l\\
g_t^l\\
z_t^l\\
\end{aligned}
\right)
}
= \left(
\begin{aligned}
&sigm\\
&sigm\\
&sigm\\
&tanh\\
&hardsigm\\
\end{aligned}
\right) f_{slice}
\left(
\begin{aligned}
&s_t^{recurrent(l)} + \\
&s_t^{top-down(l)}+   \\
&s_t^{bottom-up(l)} + \\
&b_l
\end{aligned}
\right)
\end{equation}

where
\begin{equation}\label{recurrent}
s_t^{recurrent(l)} = U_l^l h_{t-1}^l
\end{equation}
\begin{equation}\label{recurrent}
s_t^{top-down(l)} = U_{l+1}^l (z_{t-1}^l \odot h_{t-1}^{l+1})
\end{equation}
\begin{equation}\label{recurrent}
s_t^{bottom-up(l)} = W_{l-1}^l (z_{t}^{l-1} \odot h_{t}^{l+1})
\end{equation}
and the hardsigm is estimated using Gumbel-sigmoid which will be explained later.

\subsection{HM-AN}

The sequential problems inherent in action recognition and image captioning in computer vision can be tackled by RNN-based framework. As explained previously, HM-RNN is able to learn the hierarchical temporal structure from data and enable long-term dependencies. This inspires our proposal of HM-AN model.

On the other hand, attention has been proved very effective in action recognition \cite{sharma2015action}. In HM-AN, to capture the implicit relationships between the inputs and outputs in sequence to sequence problems, we apply both hard and soft attention mechanisms to explicitly learn the important and relevant image features regarding the specific outputs. More detailed explanation follows.

\subsubsection{Estimation of Boundary Detectors}

In the proposed HM-AN, the boundary detectors $z_t$ are estimated with Gumbel-sigmoid, which is derived directly from Gumbel-softmax proposed in \cite{jang2016categorical} and \cite{DBLP:journals/corr/MaddisonMT16}.

The Gumbel-softmax replaces the argmax in Gumbel-Max trick \cite{gumbel1954statistical}, \cite{maddison2014sampling} with following Softmax function:
\begin{equation}\label{gumbelsoftmax}
y_i = \frac{exp(log(\pi_i + g_i)/\tau)}{\sum_{j=1}^{k}exp(log(\pi_j + g_j)/\tau)}
\end{equation}
where $g_1, ..., g_k$ are $i.i.d.$ sampled from distribution Gumbel (0,1), and $\tau$ is the temperature parameter. $k$ indicates the dimension of generated Softmax vector.

To derive Gumbel-sigmoid, we firstly re-write Sigmoid function as a Softmax of two variables: $\pi_i$ and 0.

\begin{equation}\label{sigmoid}
\begin{aligned}
sigm(\pi_i)& =\frac{1}{(1 + exp({ - \pi_i}))} =\frac{1}{(1 + exp({0 - \pi_i}))}   \\
& =  \frac{1}{1 + {exp(0)}/{exp(\pi_i)}}  =\frac{exp({\pi_i})}{(exp({\pi_i}) + exp(0))}
\end{aligned}
\end{equation}

Hence, the Gumbel-sigmoid can be written as:

\begin{equation}\label{gumbelsigmoid}
y_i = \frac{exp(log(\pi_i+g_i/{\tau})} {exp({log(\pi_i+g_i)/{\tau}})+exp({log(g^\prime)/{\tau}})}
\end{equation}
where $g_i, g^\prime$ are independently sampled from distribution Gumbel (0,1).

To obtain a discrete value, we set values of $z_t = \widetilde{y_i}$ as:

\begin{equation}\label{gumbelsigmoid2}
\widetilde{y_i} = \begin{cases}
1 \qquad   y_i \geq 0.5 \\
0 \qquad   otherwise

\end{cases}
\end{equation}

In our experiments, all the boundary detectors $z_t$ are estimated using the Gumbel-sigmoid with a constant temperature 0.3.

\subsubsection{Deterministic Soft Attention}
\begin{figure*}[t]
  \centering
  \includegraphics[width=\textwidth]{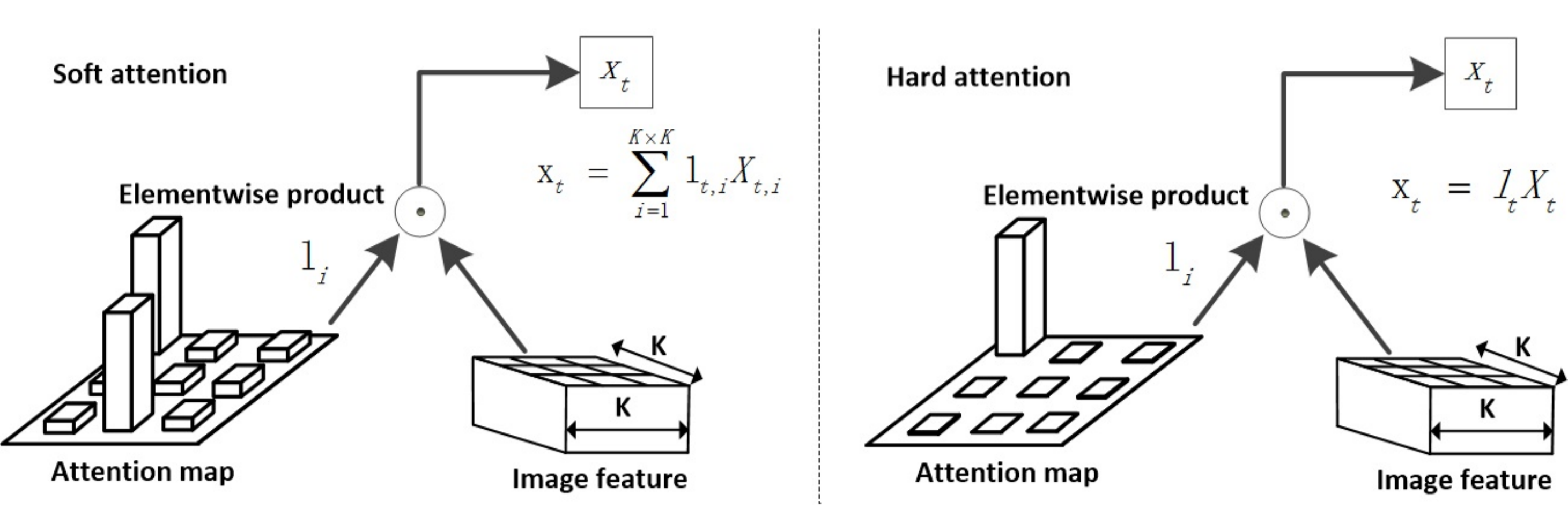}\\
  \caption{The attention mechanism: The soft attention assign weights on different locations of features using softmax whilst the values of hard attention map are either 1 or 0 which means only one important location is selected. }\label{attention}
\end{figure*}

To implement soft attention over image regions for action recognition task, we applied a similar strategy with the soft attention mechanism in \cite{sharma2015action} and \cite{xu2015show}.

Specifically, the model predicts a Softmax over K$\times$K image locations. The location Softmax is defined as:

\begin{equation}\label{deterministic}
l_{t,i} = \frac{exp(W_i h_{t-1})}{\sum_{j=1}^{K \times K}exp(W_j h_{t-1})}  \qquad i = 1, ..., K^2
\end{equation}
where i means the ith location corresponding to the specific regions in the original image.

This Softmax can be considered as the probability with which the model learns the specific regions in the image, which is important for the task at hand. Once these probabilities are obtained, the model computes the expected value of inputs by taking expectation over image features at different regions:

\begin{equation}\label{deterministic2}
x_t = \sum_{i=1}^{K^2}l_{t,i}X_{t,i}
\end{equation}
where $x_t$ is considered as inputs of the HM-AN networks. In our HM-AN implementations, the hidden states used to determine the region softmax is only defined for the first layer, i.e., $h_{t-1}^1$. The upper layers will automatically learn the abstract information of input features as explained previously. The soft attention mechanism can be visualized in the left side of Fig. \ref{attention}.

\subsubsection{Stochastic Hard Attention}

\paragraph{REINFORCE-like algorithm}\label{Reinforce_describe}
The stochastic hard attention was proposed in \cite{xu2015show}. Their hard attention was realized with the aid of REINFORCE-like algorithm. In this section, we also introduce this kind of hard attention mechanism.

The location variable $l_t$ indicates where the model decides to focus attention on the $t^{th}$ frame of a video. $l_{t,i}$ is an indicator of one-hot representation which can be set to 1 if the $i^{th}$ location contains relevant feature.

Specifically, we assign a hard attentive location of $\{\alpha_i\}$:
\begin{equation}\label{reinforce1}
\begin{aligned}
p(l_{i,t} = 1 | l_{j<t,a})& = argmax(\alpha_{t,i}) \\
&= argmax \left( \frac{exp(W_i h_{t-1})}{\sum_{j=1}^{K \times K}exp(W_j h_{t-1})} \right)
\end{aligned}
\end{equation}
where $a$ represents the input image features.

We can define an objective function $L_l$ that is a variational lower bound on the marginal log-likelihood $log\ p(y|a)$ of observing the action label $y$ given image features $a$. Hence, $L_l$ can be represented as:
\begin{equation}\label{reinforce3}
\begin{aligned}
L_l & = \sum_lp(l|a)log\ p(y|l,a) \\
   & \leq log\ \sum_lp(l|a)p(y|l,a)  \\
   & = log p(y|a)
\end{aligned}
\end{equation}

\begin{equation}\label{reinforce4}
\begin{aligned}
\frac{\partial L_l}{\partial W} = \sum_l p(l|a)
[ \frac{\partial log\ p(y|l,a)}{\partial W} + \\
log\ p(y|l,a) \frac{\partial log\ p(l|a)}{\partial W}]
\end{aligned}
\end{equation}

Ideally, we would like to compute the gradients of Equation \ref{reinforce4}. However, it is not feasible to compute the gradient of expectation in Equation \ref{reinforce4}. Hence, a Monte Carlo approximation technique is applied to estimate the gradient of the operation of expectation.

Hence, the derivatives of the objective function with respect to the network parameters can be expressed as:

\begin{equation}\label{reinforce5}
\begin{aligned}
\frac{\partial L_l}{\partial W} = \frac{1}{N}\sum_{n=1}^N
[ \frac{\partial log\ p(y|\tilde{l}_n,a)}{\partial W} + \\
log\ p(y|\tilde{l}_n,a) \frac{\partial log\ p(\tilde{l}_n|a)}{\partial W}]
\end{aligned}
\end{equation}
where $\tilde{l}$ is obtained based on the argmax operation as in Equation \ref{reinforce1}.

Similar with the approaches in \cite{xu2015show}, a variance reduction technique is used.  With the $k^{th}$ mini-batch, the moving average baseline is estimated as an accumulation of the previous log-likelihoods with exponential decay:

\begin{equation}\label{reinforce6}
b_k = 0.9 \times b_{k-1} + 0.1 \times log\ p(y|\tilde{l}_k,a)
\end{equation}

The learning rule for this hard attention mechanism is defined as follows:
\begin{equation}\label{reinforce7}
\begin{aligned}
\frac{\partial L_l}{\partial W}  \approx \frac{1}{N}\sum_{n=1}^N
[ \frac{\partial log\ p(y|\tilde{l}_n,a)}{\partial W} + \\
\lambda (log\ p(y|\tilde{l}_n,a) -b) \frac{\partial log\ p(\tilde{l}_n|a)}{\partial W}]
\end{aligned}
\end{equation}
where $\lambda$ is a pre-defined parameter.

As being pointed out in Ba et al. \cite{ba2014multiple}, Mnih et al. \cite{mnih2014recurrent} and Xu et al. \cite{xu2015show}, this is a formulation which is equivalent to the REINFORCE learning rule \cite{williams1992simple}. For convenience, it is abbreviated as REINFORCE-Hard Attention in the following.

\paragraph{Gumbel Softmax}
In hard attention mechanism, the model selects one important region instead of taking the expectation. Hence, it is a stochastic unit which can not be trained using back propagation. \cite{xu2015show} applied the REINFORCE to estimate the gradient of the stochastic neuron. Although the REINFORCE is an unbiased estimator, the variance of gradient is large and the algorithm is complex to implement. To solve these problems, we propose to apply Gumbel-softmax to estimate the gradient of the discrete units in our model. Gumbel-softmax is better in practice \cite{jang2016categorical} and much easier to implement.

We can simply replace the Softmax with Gumbel-softmax in Equation \ref{deterministic} and remove the process of taking expectation to realize the hard attention.

\begin{equation}\label{stochastic}
l_{t,i} = \frac{exp(log(W_i h_{t-1}+ g_i)/\tau)}{\sum_{j=1}^{K \times K}exp(log(W_j h_{t-1}+g_j)/\tau)}  \qquad i = 1 ... K^2
\end{equation}

The Gumbel-softmax will choose a single location indicating the most important image region for the task. However, the searching space for the parameter of temperature is too large to be manually selected. And the temperature is a sensitive parameter as explained in \cite{jang2016categorical}. Hence in this paper we applied the adaptive temperature as in \cite{gulcehre2017memory}. The adaptive temperature is to determine the value of temperature adaptively depending on current hidden states. In other words, instead of being treated as pre-defined parameter, the value of temperature is learnt from data. Specifically, we can set the following mechanism to determine the temperature:

\begin{equation}\label{temperature}
\tau = \frac{1}{Softplus(W_{temp} h_t^1+ b_{temp}) + 1}
\end{equation}
where $h_t^1$ is the hidden state of first layer of our HM-AN. Equation \ref{temperature} generates a scalar for temperature. In the equation, adding 1 can enable the temperature fall into the scope of 0 and 1. The hard attention mechanism can be seen in the right side of Fig. \ref{attention}.

\subsection{Application of HM-AN in Action Recognition}

The proposed HM-AN can be directly applied in video action recognition. In video action recognition, the dynamics exist in the inputs, i.e., the given video frames. With attention mechanism embedded in RNN, the important features of each frames can be discovered and discriminated in order to facilitate the recognition.

\begin{figure*}[t]
  \centering
  \includegraphics[width=\textwidth]{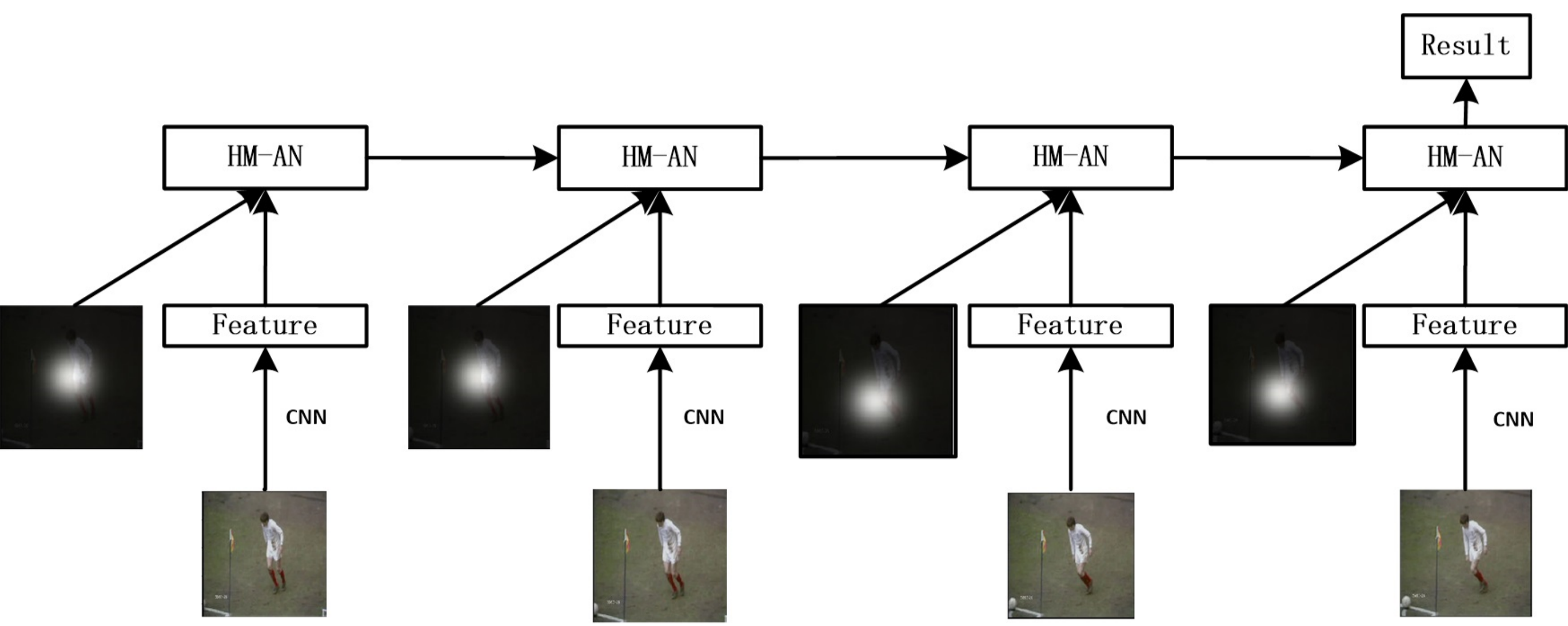}\\
  \caption{Action recognition with HM-AN.}\label{action}
\end{figure*}

For action recognition, the HM-AN applies the cross-entropy loss for recognition.
\begin{equation}\label{Cost}
LOSS = -\sum\limits_{t=1}^T\sum\limits_{i=1}^Cy_{t,i}log(\hat{y}_{t,i})
\end{equation}
where $y_t$ is the label vector, $\hat{y}_{t}$ is the classification probabilities at time step t. $T$ is the number of time steps and $C$ is the number of action categories.
The system architecture of action recognition using HM-AN is shown in Fig. \ref{action}

\section{Experiments}

In this section, we first explain our implementation details then report the experimental results on action recognition.

\subsection{Implementation Details}

We implemented the HM-AN in the platform of Theano \cite{bastien2012theano} and all the experiments were conducted on a server embedded with a Titan X GPU.  In our experiments, the HM-AN is a three layer stacked RNN. The outputs are concatenated by hidden states from three layers and forwarded to a softmax layer.

In addition to the baseline approach (LSTM networks), 4 versions of HM-AN were implemented for the purpose of comparison:
\begin{itemize}
  \item LSTM with soft attention (Baseline). The baseline approach is set as LSTM networks with soft attention mechanism.
  \item Deterministic soft attention in HM-AN (Soft Attention). This is to see how soft attention mechanism performs with the HM-AN.
  \item Stochastic hard attention with reinforcement learning in HM-AN (REINFORCE-Hard Attention). This kind of hard attention mechanism is described in Section \ref{Reinforce_describe}.
  \item Stochastic hard attention with 0.3 temperature for Gumbel-softmax in HM-AN (Constant-Gumbel-Hard Attention). The constant temperature is applied in Gumbel-softmax to accomplish the proposed hard attention model.
  \item Stochastic hard attention with adaptive temperature for Gumbel-softmax in HM-AN (Adaptive-Gumbel-Hard Attention). The temperature is set as a function of hidden states of RNN.
\end{itemize}

For the experiments, we firstly extracted frame-level CNN features using MatConvNet \cite{vedaldi2015matconvnet} based on Residue-152 Networks \cite{he2015deep} trained on the ImageNet \cite{deng2009imagenet} dataset. The images were resized to 224$\times$224, hence the dimension of each frame-level features is 7$\times$7$\times$2048. For the network training, we applied mini-batch size of 64 samples at each iteration. For each video clip, the baseline approach randomly selected 30 frames for training while the proposed approaches selected 60 frames for training as the proposed HM-AN is able to capture long-term dependencies. Actually, the optimal length for LSTM with attention is 30 and increasing the number will seriously deteriorate the performance. We applied the back propagation algorithm through time and Adam optimizer \cite{kingma2014adam} with a learning rate of 0.0001 to train the networks. The learning rate was changed to 0.00001 after 10,000 iterations. At test time, we compute class predictions for each time step and then average
those predictions over 60 frames. To obtain a prediction for the entire video clip, we average the predictions from all 60 frame blocks in the video. Also, we find normally the networks converge in several epoches.

\subsection{Experimental Results and Analysis}

\subsubsection{Datasets}

\begin{figure}[!t]
  \centering
  \subfigure[UCF sports dataset]{
  \includegraphics[width=\linewidth]{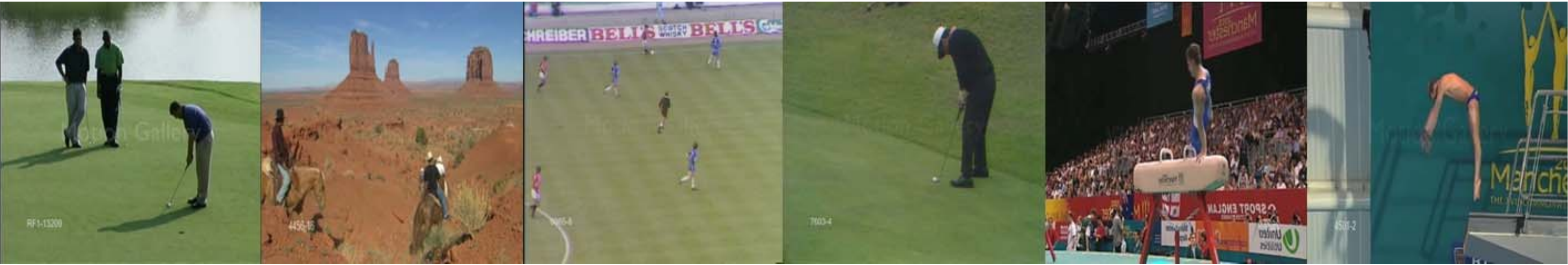}
  }

   \subfigure[Olympic sports dataset]{
  \includegraphics[width=\linewidth]{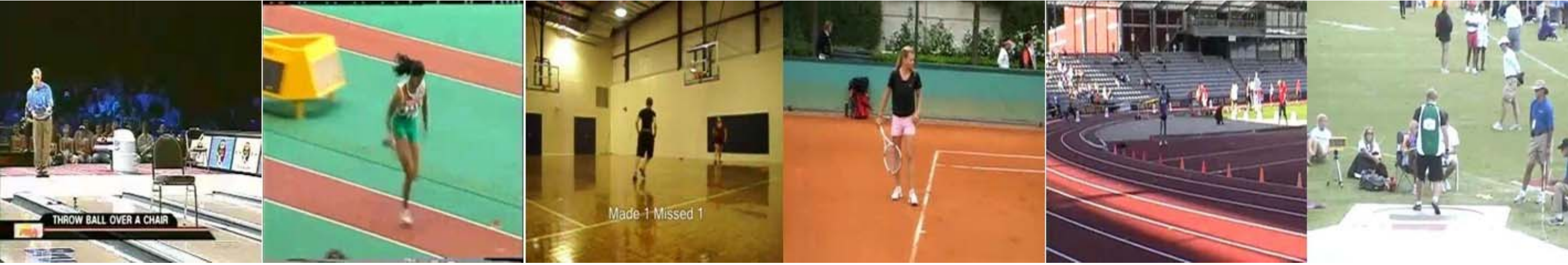}
  }

  \subfigure[HMDB51 dataset]{
  \includegraphics[width=\linewidth]{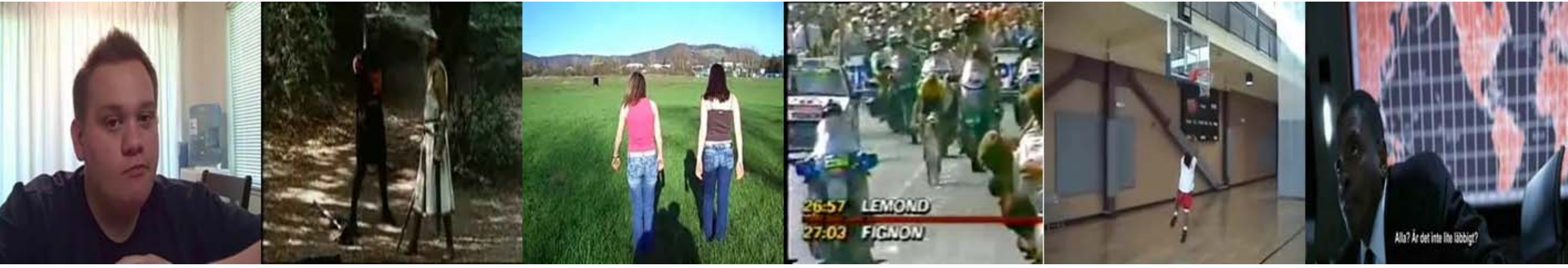}
  }

\caption{Some examples from the datasets used in this paper.}\label{dataset}
\end{figure}

We evaluated our approach on three widely used datasets, namely UCF sports \cite{rodriguez2010spatio}, Olympic sports datasets \cite{niebles2010modeling} and a more difficult Human Motion Database (HMDB51) dataset \cite{Kuehne11}. Fig. \ref{dataset} provides some examples of the three datasets used in this paper. UCF sports dataset contains a set of actions collected from various sports which are typically featured on broadcast channels such as ESPN or BBC. This dataset consists of 150 videos with the resolution of 720 $\times$ 480 and with 10 different action categories in it. Olympic sports dataset was collected from YouTube sequences \cite{niebles2010modeling} and contains 16 different sports categories with 50 videos per class. Hence, there are total 800 videos in this dataset. The HMDB51 dataset is a more difficult dataset which provides three train-test splits each consisting of 5100 videos. This clips are labeled with 51 action categories. The training set for each split has 3570 videos and the test set has 1530 videos.

For UCF sports dataset, as there is lack of training-testing split for evaluation, we manually divide the dataset into training and testing sets. We randomly pick up 75 percent for training, and leave the remaining 25 percent for testing. We then report the classification accuracy on the testing dataset.

As for Olympic sports dataset, we use the original training-testing split with 649 clips for training and 134 clips for testing provided in dataset. Following the practice in \cite{niebles2010modeling}, we evaluate the Average Precision (AP) for each category on this dataset.

When evaluating our method on HMDB51, we also follow the original training-testing split and report the classification accuracy of split 1 on the testing set.

 \begin{figure}[!t]
  \centering
  \includegraphics[width=\linewidth]{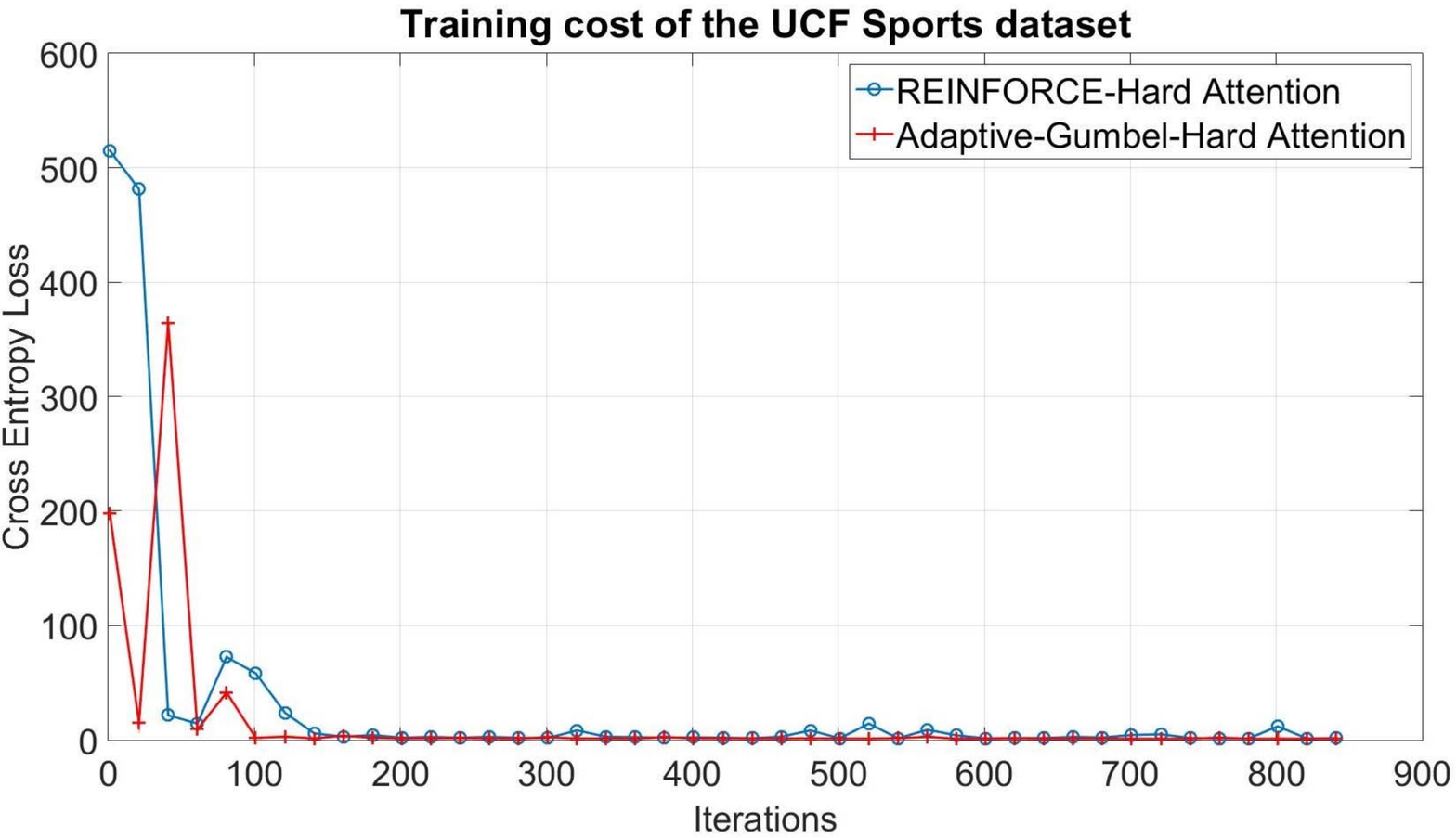}
\caption{Training cost of the UCF sports dataset.}\label{lossucf}
\end{figure}

\begin{figure}[!t]
  \centering
  \includegraphics[width=\linewidth]{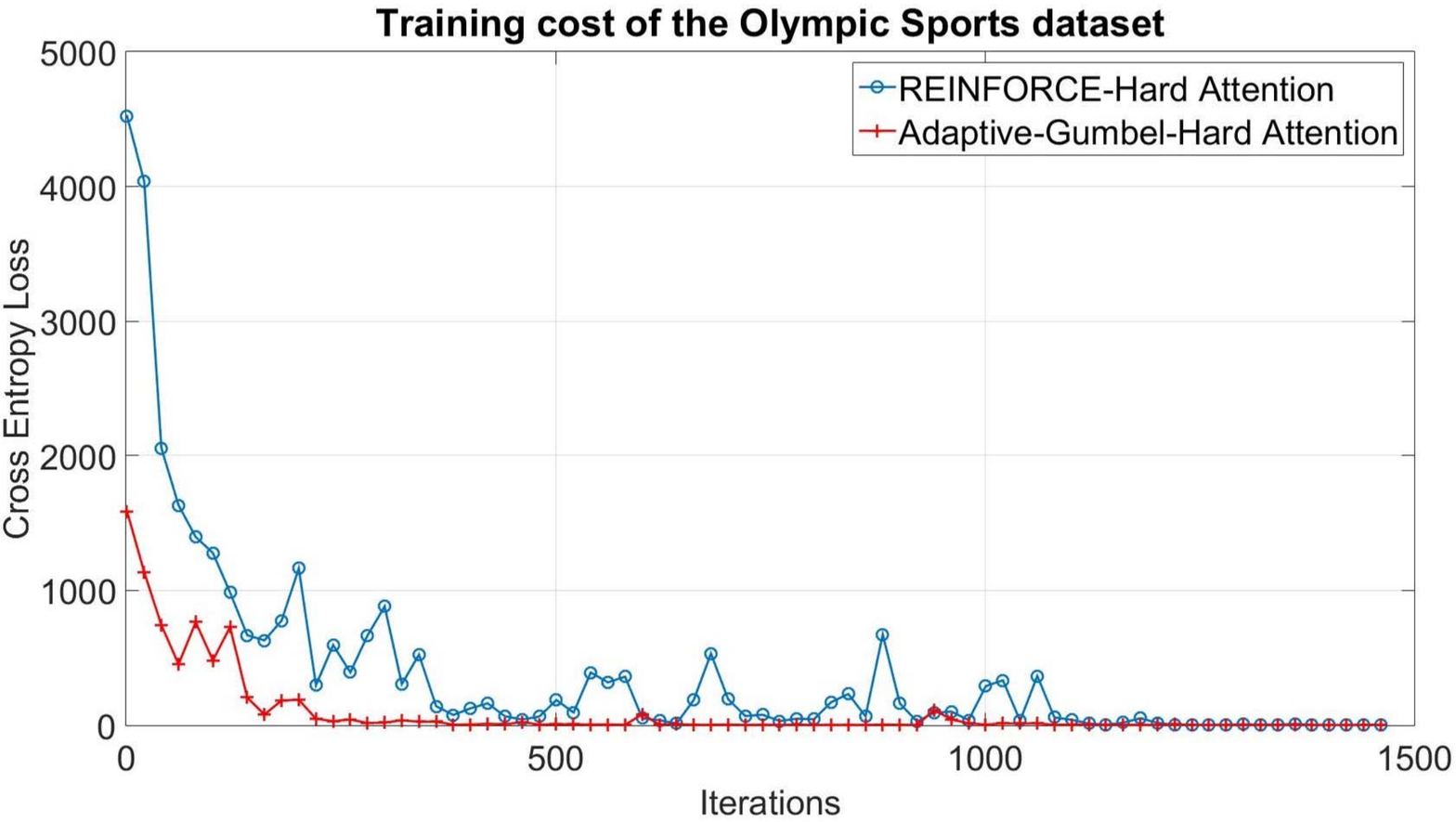}
\caption{Training cost of the Olympic sports dataset.}\label{lossolympic}
\end{figure}

\begin{figure}[!t]
  \centering
  \includegraphics[width=\linewidth]{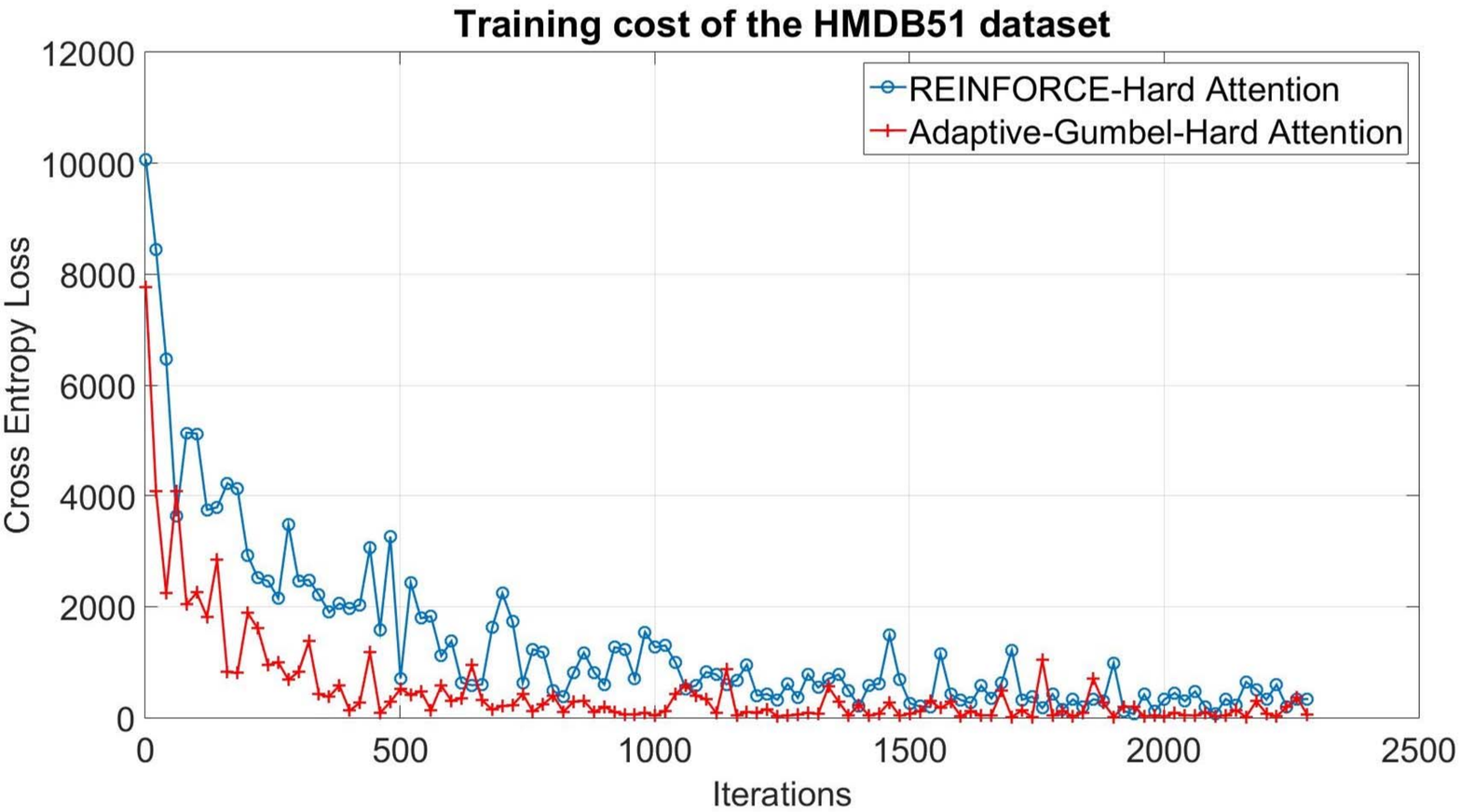}
\caption{Training cost of the HMDB51 dataset.}\label{losshmdb51}
\end{figure}

\subsubsection{Results}

\paragraph{UCF sports dataset}

We firstly tested the performance of the LSTM with soft attention proposed in \cite{sharma2015action} on the UCF sports dataset and obtained 70.0\% accuracy. All the experimental settings are the same as these in \cite{sharma2015action}. Then we evaluated the proposed four approaches mentioned previously. HM-AN with stochastic hard attention which is realized with REINFORCE-like algorithm improves the results to 82.0\%. HM-AN with soft attention is similar to the REINFORCE-Hard Attention, with accuracy results of 81.1\%. The hard attention mechanism realized by Gumbel-softmax with adaptive temperature achieves 82.0\% accuracy, similar to our REINFORCE-Hard Attention model. However, the Constant-Gumbel-Hard Attention which uses Gumbel-softmax with constant temperature value of 0.3 only yields 76.0\% accuracy, which indicates the significant role of adaptive temperature in maintaining the system performance. Fig. \ref{lossucf} shows the curves of training cost cross entropy for the Adaptive-Gumbel-Hard Attention approach and REINFORCE-Hard Attention approach, respectively. It can be seen from the figure that the REINFORCE-Hard Attention converges marginally slower than the approach of Adaptive-Gumbel-Hard Attention.

As shown in Table \ref{UCFsports}, we compare our model with the methods proposed in \cite{1705.03146} in which a convolutional LSTM attention network with hierarchical architecture was used for action recognition. The hierarchical architecture in \cite{1705.03146} was pre-defined whilst our model is able to learn the hierarchy from data. The improvements brought by our methods are obvious as shown in Table \ref{UCFsports}.

\begin{table}[!t]
\caption{Accuracy on UCF sports}
 \centering
\begin{tabular}{|c|c|}
  \hline
  Methods  & Accuracy \\
  \hline
  Baseline   \cite{sharma2015action} & 70.0\% \\
  Conv-Attention \cite{1705.03146} & 72.0\% \\
  CHAM \cite{1705.03146} & 74.0\% \\
  Soft Attention (Ours) &     81.1\%     \\
 REINFORCE-Hard Attention (Ours) & \textbf{82.0\%}\\
  Constant-Gumbel-Hard Attention (Ours)   & 76.0\% \\
  Adaptive-Gumbel-Hard Attention (Ours) & \textbf{82.0\%} \\
  \hline
\end{tabular}
\label{UCFsports}
\end{table}%

\paragraph{Olympic sports dataset}
Olympic sports dataset is of medium size. Experiments on this dataset are shown in Table \ref{Olympic}. The mAP result of baseline approach is 73.7\%. Our method HM-AN with Soft attention achieves 82.4\% mAP. However, unlike UCF sports dataset, the mAP result of REINFORCE-Hard Attention is 77.1\%, which is lower than the approach of Soft Attention. The Constant-Gumbel-Hard Attention, which is implemented by Gumbel-softmax with a constant temperature 0.3, obtains a mAP value of 82.3\%. By making the temperature value of Gumbel-softmax adaptive, the proposed model achieves 82.7\% mAP, the highest among all our experimental settings. Again, our proposed methods show superior performance compared to the hand-designed hierarchical model in \cite{1705.03146} even the convolutional LSTM with attention was utilized in \cite{1705.03146}.

\begin{figure}[!t]
  \centering
  \includegraphics[width=\linewidth]{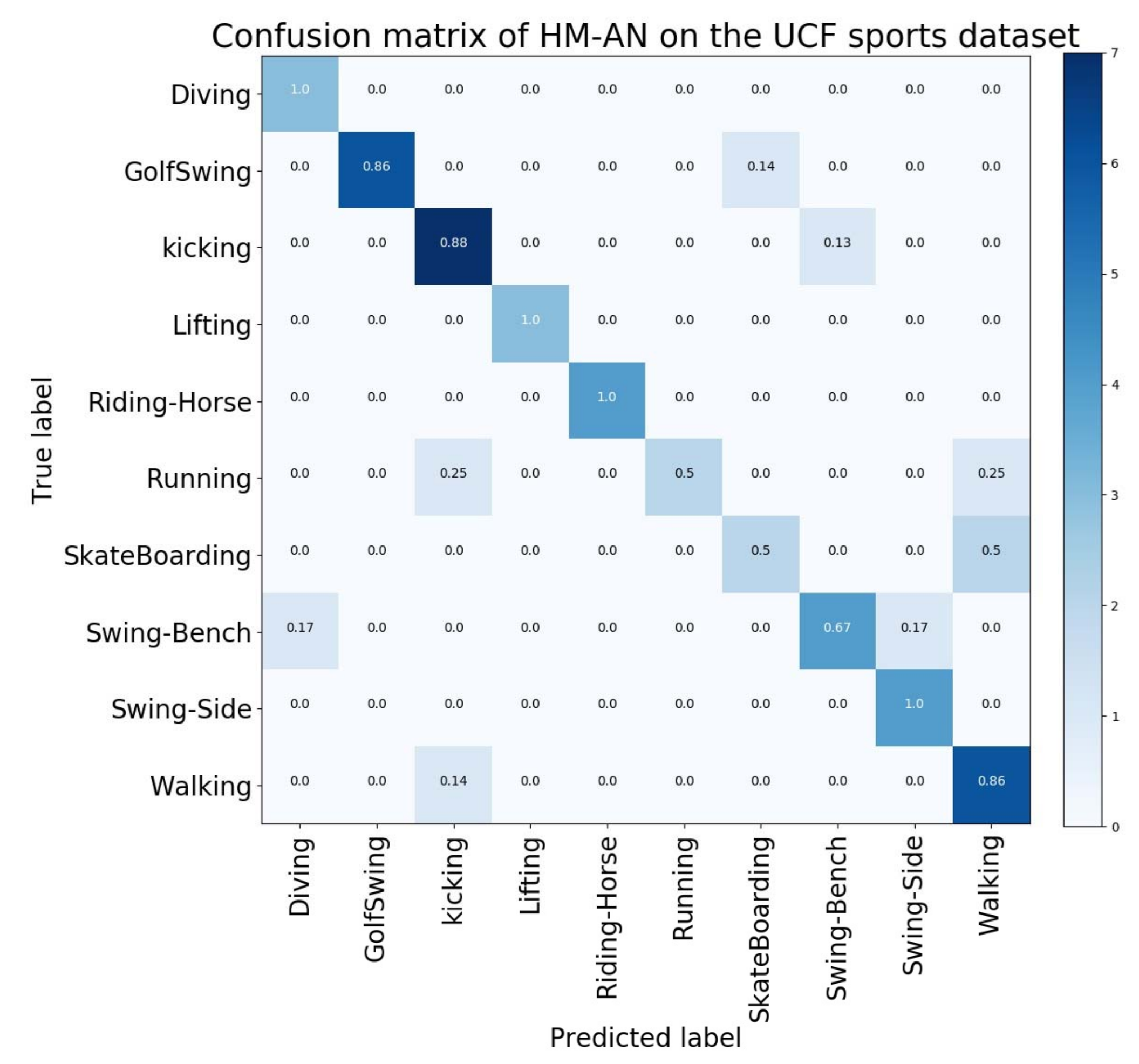}
\caption{Confusion Matrix of HM-AN with Adaptive-Gumbel-Hard Attention on the UCF sports dataset.}\label{conf_ucf}
\end{figure}

\begin{table*}[t]
\caption{AP on Olympics sports}
 \centering
 \resizebox{\textwidth}{!}{
\begin{tabular}{|c|c|c|c|c|c|c|}
  \hline
  Class & Vault &Triple Jump & Tennis serve&Spring board     &   Snatch\\
\hline
   Baseline \cite{sharma2015action} & 97.0\% &88.4\%  &52.3\%  &60.0\% &23.2\%  \\

   Conv-Attention \cite{1705.03146} & 97.0\% & 94.0\% & 49.8\% & 66.4\% & 26.1\% \\

   CHAM \cite{1705.03146} & 97.0\% & 98.9\% & 49.5\% & 69.2\% & 47.8\% \\

   Soft Attention(Ours)  &99.0\%  & 100.0\%    & 60.7\%  &64.2\%   & 38.6\% \\

    REINFORCE-Hard Attention (Ours) & 100.0\% & 95.0\% & 50.8\% & 56.3\% & 28.6\% \\

    Constant-Gumbel-Hard Attention (Ours)   & 97.0 \%   &99.0\%    &62.6 \% & 58.7\% &40.3\%\\

    Adaptive-Gumbel-Hard Attention (Ours) & 98.1\%   & 98.9\%    & 62.1\% & 64.3\% & 45.4\% \\
    \hline

 Shot put &Pole vault &Platform 10m & Long jump & Javelin Throw & High jump \\
 \hline
   67.4\% &69.8\%  &84.1\%  &100.0\% &89.6\% &84.4\%  \\
   60.0\% &100\%   & 86.0\% & 98.0\% & 87.9\% & 80.0\% \\
   79.8\% & 60.8\% & 89.7\% & 100.0\% & 95.0\% & 78.7\% \\
   77.2\%  & 85.4\% & 91.5\% & 98.9\% & 97.0 & 77.2\%   \\
    90.6\%  &100.0\%  &86.7\% &     100.0\%   &  89.7\%  &  77.5\%      \\
    87.8\% & 100.0\% & 93.1\% &    100.0\%  &93.2\% & 82.8\%  \\
    84.1\% & 100.0\% & 94.8\% &     100.0\% & 95.3\% & 86.2\%    \\

    \hline

  Hammer throw  &Discus throw &Clean and jerk &Bowling &Basketball layup & mAP\\
\hline
    38.0\% & 100.0\% &76.0\% &60.0\%  &89.8\%  &  73.7\%    \\
    36.6\% & 97.8\%& 100.0\% & 46.8\% & 81.2\% & 75.5\%      \\
    37.9\% & 97.0\% & 84.8\% & 46.7\% & 89.1\% & 76.4\%       \\
    44.1\% & 94.2\% & 83.8\% & 63.9\% & 89.2\% & 77.1\%       \\
    52.9\%  &95.8\%  & 92.4\%  & 69.4\% &  98.1\% &  82.4\%    \\
     54.7\%  & 95.8\%  &  91.3\% & 60.5\%   & 100.0\%    &   82.3\%  \\
     53.8\%  &95.8\%  &84.9\% & 62.5\%  &  97.0\%   &   \textbf{82.7\%}  \\
  \hline
\end{tabular}
}
\label{Olympic}
\end{table*}%

\paragraph{HMDB51 dataset}

HMDB51 is a more difficult and bigger dataset. The performance of baseline approach is reported in \cite{sharma2015action}, with 41.3\% accuracy. Our HM-AN model with soft attention improves the accuracy to 43.8\%. We then apply the REINFORCE-Hard Attention approach on this dataset. The accuracy result turns out to be lower than the HM-AN with soft attention. Moreover, the model with REINFORCE-like algorithm converges slower than the Gumbel-softmax with adaptive temperature, also with more oscillations on the training cost, which is shown in Fig. \ref{losshmdb51}.  With constant value 0.3 of temperature for hard attention, the model achieves 44.0\% accuracy. Again, the improvement by adding adaptive temperature is obvious, with 44.2\% accuracy on HMDB51 dataset. The accuracy results are further summarized in Table \ref{HMDB51}. We also compare the performance of the proposed HM-AN with some published models related to ours. Firstly, the methods not from RNN family but only with the spatial image shows poor performance as shown in Table \ref{HMDB51Cmp}, even with CNN model fine-tuned. Specifically, the softmax regression results from \cite{sharma2015action} directly extracted image feature of each frame and perform softmax regression on them, with 33.5\% accuracy. The spatial convolutional net from the famous two-stream approach \cite{simonyan2014two} can be considered as a similar method with the difference that the two-stream approach performs fine-tuning on the CNN model, with an improved accuracy of 40.5\%. The LSTM without attention also achieves 40.5\% accuracy \cite{sharma2015action}. When adding soft attention mechanism, an improved accuracy of 41.3\% can be obtained. The Conv-Attention \cite{1705.03146} and ConvALSTM \cite{li2016videolstm} both use convolutional LSTM with attention. The differences are that Conv-Attention extracts features from Residual-152 Networks \cite{he2016deep} without fine-tuning whilst ConvALSTM extracts image features from a fine-tuned VGG16 model. The ConvALSTM leads Conv-Attention by a small margin, with 43.3\% accuracy. As explained previously, CHAM \cite{1705.03146} has a hand-designed hierarchical architecture in contrast with ours in which the temporal hierarchy is formed through training. Our best setting (Adaptive-Gumbel-Hard Attention) reports the highest accuracy (44.2\%) among methods from the RNN family and leads the CHAM results (43.4\%) by 0.8 percent.

\begin{table}[!t]
\caption{Accuracy on HMDB51}
 \centering

\begin{tabular}{|c|c|}
  \hline
  Methods  & Accuracy \\
  \hline
  Baseline   \cite{sharma2015action} & 41.3\% \\
  Soft Attention (Ours)&     43.8\%     \\
  REINFORCE-Hard Attention(Ours) & 41.5\% \\
  Constant-Gumbel-Hard Attention (Ours)   & 44.0\% \\
  Adaptive-Gumbel-Hard Attention (Ours) &  \textbf{44.2\%}  \\
  \hline
\end{tabular}

\label{HMDB51}
\end{table}%

\begin{table*}[!t]
\caption{Comparison with related methods on HMDB51}
 \centering
   \resizebox{\textwidth}{!}{
\begin{tabular}{|c|c|c|c|}
  \hline
  Methods  & Accuracy & Spatial Image Only & Fine-tuning of CNN model \\
  \hline
  Softmax Rgression \cite{sharma2015action} & 33.5\% & Yes & No \\
  Spatial Convolutional Net \cite{simonyan2014two} & 40.5\% & Yes & Yes \\
  Trajectory-based modeling \cite{jiang2012trajectory} & 40.7\% & No & No \\
  \hline
  Average pooled LSTM \cite{sharma2015action} & 40.5\% & Yes & No \\
  Baseline   \cite{sharma2015action} & 41.3\% & Yes & No \\
  Conv-Attention \cite{1705.03146} & 42.2\% & Yes & No \\
  ConvALSTM \cite{li2016videolstm} & 43.3\% & Yes & Yes \\
  CHAM \cite{1705.03146} & 43.4\% & Yes & No \\
  Adaptive-Gumbel-Hard Attention (Ours)   & \textbf{44.2\%}& Yes & No \\
  \hline
\end{tabular}
}
\label{HMDB51Cmp}
\end{table*}%

\begin{figure*}[!t]
  \centering
  \includegraphics[width=\textwidth]{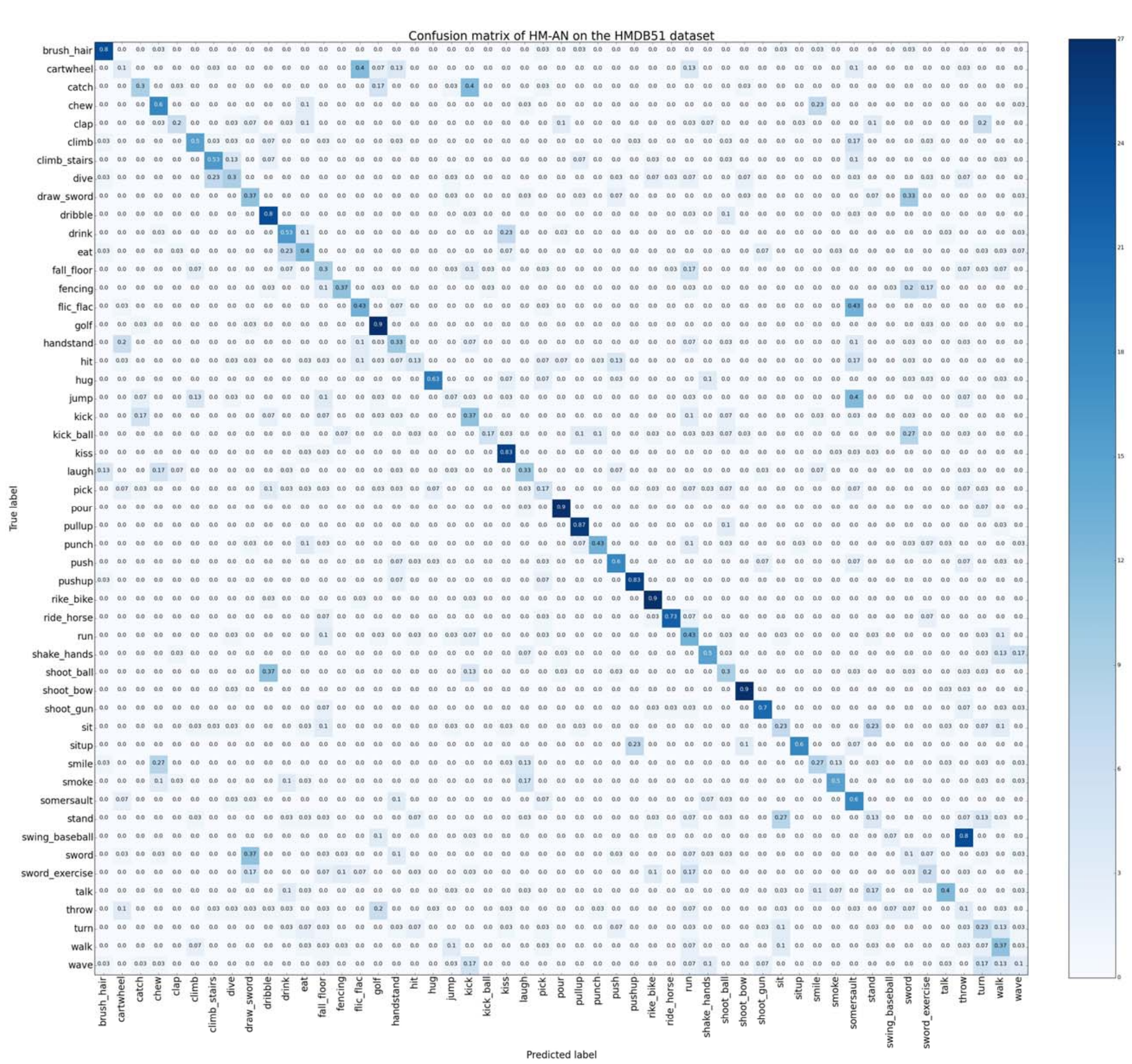}
\caption{Confusion Matrix of HM-AN Adaptive-Gumbel-Hard Attention on the HMDB51 dataset.}\label{conf_hmdb51}
\end{figure*}

\paragraph{Analysis and Visualization}

We tested four approaches (Soft Attention, REINFORCE-Hard Attention, Constant-Gumbel-Hard Attention and Adaptive-Gumbel-Hard Attention) on three different datasets: UCF sports dataset, Olympic sports dataset and HMDB51 dataset. On UCF sports dataset, The REINFORCE-Hard Attention and Adaptive-Gumbel-Hard Attention generate satisfactory results and shows better performance than the soft attention and Constant-Gumbel-Hard Attention. This indicates that the adaptive temperature is an efficient method to improve performance in implementation of Gumbel-softmax based hard attention.

On both of the Olympic sports dataset and HMDB51 dataset, the best approach is the Adaptive-Gumbel-Hard Attention while the REINFORCE-Hard Attention is even worse than soft attention mechanism. On the bigger datasets, the advantages of Gumbel-softmax include small gradient variance and simplicity, which are obvious compared with the REINFORCE-like algorithms. This shows that Gumbel-softmax generalizes well on large and complex datasets. This is reflected not only by the accuracy results, but also by the training cost curves in Fig. \ref{lossolympic} and Fig. \ref{losshmdb51}. This conclusion is also consistent with the findings in other recent researches \cite{gulcehre2017memory} which also applied both REINFORCE-like algorithms and Gumbel-softmax as estimators for stochastic neurons.

\begin{figure*}[!t]
  \centering
  \includegraphics[width=\linewidth]{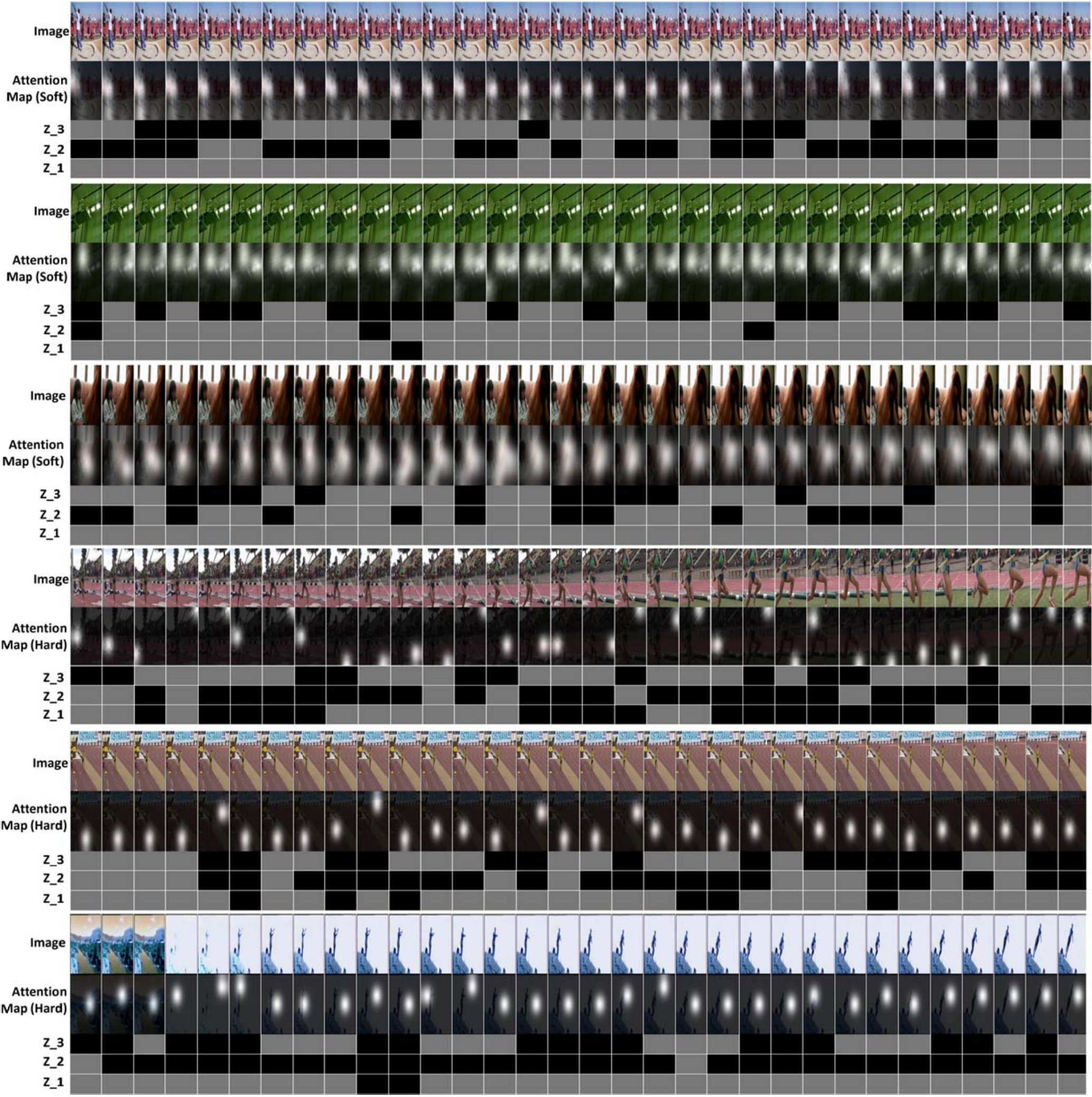}
\caption{Visualization of attention maps and detected boundaries for action recognition.}\label{actionvis}
\end{figure*}

The visualization of attention maps and boundary detectors learnt by the HM-AN is given in Fig. \ref{actionvis}. In the attention maps, the brighter an area is, the more important it is for the recognition. The soft attention captures multi-regions while the hard attention selects only one important region. As can be seen from the figure, in different time steps, the attention regions are different which means the model is able to select region to facilitate the recognition through time automatically. The $z\_1$, $z\_2$ and $z\_3$ in the figure indicate the boundary detectors in the first layer, the second layer and the third layer, respectively. In the figure for boundary detector, the black regions indicate there exists a boundary in the time-domain whilst the grey regions show the UPDATE operation can be performed. The multi-scale properties in the time-domain can be captured by the HM-AN as different layer shows different boundaries.

\section{Conclusion}

In this paper, we proposed a novel RNN model, HM-AN, which improves HM-RNN with attention mechanism for visual tasks. Specifically, the boundary detectors in HM-AN is implemented by recently proposed Gumbel-sigmoid. Two versions of attention mechanism are implemented and tested. Our work is the first attempt to implement hard attention in vision task with the aid of Gumbel-softmax instead of REINFORCE algorithm. To solve the problem of sensitive parameter of softmax temperature, we applied the adaptive temperature methods to improve the system performance. To validate the effectiveness of HM-AN, we conducted experiments on action recognition from videos. Through experimenting, we showed that HM-AN is more effective than LSTMs with attention. The attention regions of both hard and soft attention and boundaries detected in the networks provide visualization for the insights of what the networks have learnt. Theoretically, our model can be built based on various features, e.g., Dense Trajectory, to further improve the performance. However, our emphasis in this paper is to prove the superiority of the model itself compare with other RNN-like model given same features. Hence, we chose to use deep spatial features only. Our work would facilitate further research on the hierarchical RNNs and its applications on the vision tasks.


%

\bibliographystyle{IEEEtran}
\bibliography{bib}

\end{document}